\documentclass[11pt]{llncs}
\usepackage[margin=1in]{geometry}
\usepackage{hyperref}

\usepackage{amssymb, amsfonts, amsmath}
\usepackage [autostyle, english = american]{csquotes}
\usepackage{tabularx}
\usepackage{graphicx}
\usepackage{url}
\usepackage{bbm}
\usepackage{caption} 
\usepackage[cal=boondox]{mathalfa}

\MakeOuterQuote{"}

\newcommand{\size}[1]{\vert #1 \vert}

\newcommand{\figref}[1]{Figure~\ref{fig:#1}}
\newcommand{\tabref}[1]{Table~\ref{tab:#1}}
\renewcommand{\eqref}[1]{Equation~\ref{eq:#1}}
\newcommand{\eqsthreeref}[3]{Equations~\ref{eq:#1}, ~\ref{eq:#2}, and~\ref{eq:#3}}

\newcommand{\secref}[1]{Section~\ref{sec:#1}}
\newcommand{\appref}[1]{Appendix~\ref{app:#1}}
\renewcommand{\eqref}[1]{Equation~\ref{eq:#1}}
\newcommand{\thmref}[1]{Theorem~\ref{thm:#1}}
\newcommand{\lemref}[1]{Lemma~\ref{lem:#1}}

\newcommand{\BigO}[1]{\ensuremath{\operatorname{O}\!\left(#1\right)}}
\newcommand{\LittleO}[1]{\ensuremath{\operatorname{o}\!\left(#1\right)}}

\usepackage{natbib}
\newcommand\newcite{\citet}
\renewcommand\cite{\citep}
\begin{document}

\title{Optimal Subarchitecture Extraction For BERT}

\author{Adrian de Wynter \and %
Daniel J. Perry}%

\authorrunning{A. de Wynter and D. J. Perry}

\institute{$^1$Amazon Alexa, 300 Pine St., Seattle, Washington, USA 98101 \\
  \email{\{dwynter,prrdani\}@amazon.com}
}

\maketitle

\begin{abstract}\footnote{This paper has not been fully peer-reviewed.}
We extract an optimal subset of architectural parameters for the BERT architecture from \newcite{BERT} by applying recent breakthroughs in algorithms for neural architecture search. 
This optimal subset, which we refer to as "Bort", is demonstrably smaller, having an effective (that is, not counting the embedding layer) size of $5.5\%$ the original BERT-large architecture, and $16\%$ of the net size. 
Bort is also able to be pretrained in $288$ GPU hours, which is $1.2\%$ of the time required to pretrain the highest-performing BERT parametric architectural variant, RoBERTa-large \cite{RoBERTa}, and about $33\%$ of that of the world-record, in GPU hours, required to train BERT-large on the same hardware.\footnote{\url{https://developer.nvidia.com/blog/training-bert-with-gpus/}, accessed July $30^{th}, 2020$.} 
It is also $7.9$x faster on a CPU, as well as being better performing than other compressed variants of the architecture, and some of the non-compressed variants: it obtains performance improvements of between $0.3\%$ and $31\%$, absolute, with respect to BERT-large, on multiple public natural language understanding (NLU) benchmarks. 
\end{abstract}

\section{Introduction}

Ever since its introduction by \newcite{BERT}, the BERT architecture has had a resounding impact on the way contemporary language modeling is carried out. Such success can be attributed to this model's incredible performance across a myriad of tasks, solely requiring the addition of a single-layer linear classifier, and a simple fine-tuning strategy. 
On the other hand, BERT's usability is considered an issue for many, due to its large size, slow inference time, and complex pre-training process. Many attempts have been done to extract a simpler subarchitecture that maintains the same performance of its predecessor, while simplifying the pre-training process--as well as the inference time--to varying degrees of success. 
Yet, the performance of such subarchitectures is still being surpassed by the original implementation in terms of accuracy, and the choice of the set of architectural parameters in these works often appears to be arbitrary.

While this problem is hard in the computational sense, recent work by \newcite{architecturesearch} suggests that there exists an approximation algorithm--more specifically, a fully polynomial-time approximation scheme, or FPTAS--that, under certain conditions, is able to efficiently extract such a set with optimal guarantees. 

\subsection{Contributions}
We consider the problem of extracting the set of architectural parameters for BERT that is optimal over three metrics: inference latency, parameter size, and error rate. We prove that BERT presents the set of conditions, known as the \emph{strong} $AB^nC$ \emph{property}, required for the aforementioned algorithm to behave like an FPTAS. We then extract an optimal subarchitecture from a high-performing BERT variant, which we refer to in this paper as "Bort". This optimal subarchitecture is $16\%$ the size of BERT-large, and performs inference eight times faster on a CPU.

Although the FPTAS guarantees which architecture will perform best--rather, it returns the architectural parameter set optimal over the three metrics being considered--it does not yield an architecture trained to convergence. 
Accordingly, we pre-train Bort and find that the pre-training time is remarkably improved with respect to its original counterpart: $288$ GPU hours versus $1,\!153$ for BERT-large and $24,\!576$ for RoBERTa-large, on the same hardware and with a dataset larger than or equal to its comparands. 
We also evaluate Bort on the GLUE \cite{GLUE}, SuperGLUE \cite{SuperGLUE}, and RACE \cite{RACE} public NLU  benchmarks, obtaining significant improvements in all of them with respect to BERT-large, and ranging from $0.3\%$ to $31\%$. We release the trained model and code for the fine-tuning portion implemented in MXNet \cite{chen2015mxnet}.\footnote{The code can be found in this repository: \url{https://github.com/alexa/bort}}

\subsection{Outline}
Our paper is split in three main parts. The extraction of an optimal subarchitecture through an FPTAS, along with parameter size and inference speed comparisons is discussed in \secref{main}. Given that the FPTAS does not provide a trained architecture, we discuss the pre-training of Bort to convergence through knowledge distillation in \secref{distillation}, and contrast it with a simple, unsupervised method with respect to intrinsic evaluation metrics. The third part deals with the extrinsic analysis of Bort on various NLU benchmarks (\secref{results}). Aside from these three areas, we give an overview of related work in \secref{relatedwork}, and conclude in \secref{conclusion} with a brief discussion of our work, potential improvements to Bort, as well as interesting research directions.

\section{Related Work}\label{sec:relatedwork}

\subsection{BERT Compression}
It could be argued that finding a high-performing compressed BERT architecture has been an active area of research since the original article was released; in fact, \newcite{BERT} stated that it remained an open problem to determine what was the exact relationship amongst BERT's architectural parameters. There is no shortage of related work on this field, and significant progress has been done in the so-called "Bertology", or study of the BERT architecture. Of note is the work of \newcite{attentionheads}, who evaluate the amount of attention heads that are truly required to perform well; and \newcite{clark-etal-2019-bert} and \newcite{kovaleva-etal-2019-revealing} who analyze and prove that such architectures parse lexical and semantic features at different levels, showing predilection for certain specific tokens and other coreferent objects. 

The most prominent studies in the influence of architectural parameters in BERT's performance are the ones done with TinyBERT \cite{tinyBERT}, which comes in a $4$ and a $6$-layer variant; distillation strategies such as BERT-of-Theseus \cite{Theseus} and DistilBERT \cite{Distillbert}; as well as the recent release of a family of fully pre-trained smaller architectural variations by \newcite{GooglesMiniBERTs}. We discuss more about their contribution in \secref{relatedkdwork}. 

In the original BERT paper it was mentioned that the model was considerably underfit. The work by \newcite{RoBERTa} addressed this by expanding the vocabulary size, as well as increasing the input sequence length, extending the training time, and changing the training objective. This architecture, named RoBERTa, is--with the possible exception of the vocabulary size--the same as BERT, yet it is the highest-performing variant accross a variety of NLU tasks. 

Of note, although not relevant to our specific study, are ALBERT \cite{ALBERT} and MobileBERT \cite{mobileBERT}, both of which are variations on the architectural graph itself--and hence not parametrizable through the FPTAS mentioned earlier. In particular, the former relies on specialized parameter sharing, in addition to the reparametrization of the architecture with which we deal on this work. Given we have constrained ourselves to solely optimize the architectural parameters, and not the graph, we do not consider either of them as a close enough relative to the BERT architecture to warrant comparison: our problem requires us to have a space of architectures parametrizable in a precise sense, as described in \secref{fptasexplanation}. Still, it is worth mentioning that ALBERT has shown outstanding results in an ensemble setting across multiple NLU tasks, and MobileBERT is remarkably high performing given its parameter size.

\subsection{Knowledge Distillation}\label{sec:relatedkdwork}

Knowledge distillation (KD) was, to our knowledge, originally proposed by \newcite{bucilu2006model} primarily as a model compression scheme. 
This was further developed by \newcite{ba2014deep} and \newcite{romero2014fitnets}, and brought to mainstream by \newcite{hinton2015distilling} where they explored additional techniques such as temperature modification of each logit output, and the use of unlabeled transfer sets. 
Some of the earliest examples of this approach were first used in image classification and automatic speech recognition \cite{hinton2015distilling}; however, more recently, KD has also been used to compress language models. 

The work done by \newcite{Distillbert} and \newcite{sun2019patient}, for example, both take the approach of truncating the original BERT architecture to a smaller number of layers. This is carried out by initializing with the original weights of the larger model, and then use KD to fine-tune the resulting model. 
\newcite{tinyBERT} use a similar approach to architecture choice, but introduces the term \emph{transformer distillation} to refer to their multilayer distillation loss, similar to that introduced by \newcite{romero2014fitnets}. 
By applying distillation losses at multiple layers of the transformer they are able to achieve stronger results in the corresponding student model. 

To our knowledge, the most methodical work in architecture choice for BERT is done by \newcite{GooglesMiniBERTs}, as they choose smaller versions of BERT-large at regular depth intervals and compare them. 
They also evaluated how pre-training with and without KD compares and impacts "downstream" fine-tuned tasks. 
Our approach is similar to theirs, but rather than selecting at regular intervals and comparing, we utilize a formalized, rather than experimental, approach to extract and optimize the model's subarchitecture over different metrics.

\section{BERT's Optimal Subarchitecture}\label{sec:main}

In this section we describe the process we followed to extract Bort. We begin by establishing some common notation, and then providing a brief overlook of the the algorithm employed. Given that the quality of the approximation of said algorithm depends strongly on the input, we describe also our choice of objective functions. We conclude this section by comparing 
the parameter size and inference latencies of BERT-large, other compressed architectures, and Bort.

\subsection{Background}
The BERT architecture is a bidirectional, fully-connected transformer-based architecture. It is comprised of an embedding layer dependent on the vocabulary size ($V = 28,\!996$ tokens in the case of BERT)\footnote{The vocabulary size varies between the cased ($V = 28,\!996$) and uncased ($V = 30,\!522$) versions. We will solely refer to the cased version throughout this paper, as it is the highest-performing implementation.}, $D$ encoder layers consisting of transformers as described by \newcite{Attention}, and an output layer. Originally, the BERT architecture was released in two variations: BERT-large, with $D=24$ encoder layers, $A=16$ attention heads, $H=1,\!024$ hidden size, and $I=4,\!096$ intermediate layer size; and BERT-base, with corresponding architectural parameters $\langle D=12, A=12, H=768, I=3,\!072\rangle$. See \appref{bertmath} for a more thorough treatment of the BERT architecture from a functional perspective.

Formally, let $\Xi$ be a finite set containing valid configurations of values of the $4$-tuple $\langle D,A,H,I\rangle$--that is, the architectural parameters. In line with \newcite{architecturesearch}, we can describe the \emph{family} of BERT architectures as the codomain of some function 

\begin{equation}\label{eq:berteq}
\textsc{BERT} : \Xi \rightarrow B
\end{equation}

such that every $b \in B$ is a $D$-encoder layer variant of the architecture parametrized by a tuple $\xi  = \langle D,A,H,I \rangle \in \Xi$. That is $A$ attention heads, and trainable parameters $w_1, w_2,\dots, w_k \in W$ where every $w_i$ belongs to either one of $\mathbb{R}^{H \times I}, \mathbb{R}^{H \times H}$, or $\mathbb{R}^{I \times H}$. We sometimes will write such a variant in the form $b(X; W)$, where $X$ is an input set such that, for fixed sequence length $s$, batch size $z$, and input $x \in X$, $x \in \mathbb{N}^{z,s}$. For a formalized description of \eqref{berteq}, refer to \appref{bertmath}. 

\subsection{Algorithm}\label{sec:fptasexplanation}

We wish to find an architectural parameter set $\xi  = \langle D,A,H,I \rangle$ that optimizes three metrics: inference speed $\mathcal{i}(b(X; \cdot))$,\footnote{This quantity is, in theory, invariant of the implementation. While there is no (asymptotic) difference between the \emph{steps} and the \emph{wall-clock} time required to forward-propagate an input, we concern ourselves only with the wall-clock, on-CPU, fixed-input-size inference time, and formally state it in \secref{fptassetup}.} parameter size $\mathcal{p}(b(\cdot; W))$, and error rate $\mathcal{e}(b(X; W^{*}), Y)$, as measured relative to some set of expected values $Y$ and trained parameters $W^{*}$. We will omit the argument to the functions when there is no room for ambiguity. 

\newcite{architecturesearch} showed that, for arbitrary architectures, this is an \textbf{NP}-Hard problem. It was also shown in the same paper that, if the input and output layers are tightly bounded by the encoder layers for both $\mathcal{p}(\cdot)$ and $\mathcal{i}(\cdot)$, it is possible to solve this problem efficiently. Given that the embedding layer is a trainable lookup table and the bulk of the operations are performed in the encoder, the parameter size of said layers is large enough that is able to dominate over the three metrics, and thus it presents such a property. Additionally, the BERT architecture contains a layer normalization on every encoder layer, in addition to a GeLU \cite{GeLU} activation function. 
If the loss function used to solve this algorithm is $L$-Lipschitz smooth with respect to $W$, bounded from below, and with bounded stochastic gradients, it is said that the architectures present the strong $AB^nC$ property, and the algorithm behaves like an FPTAS, returning a $(1 - \epsilon)$-accurate solution in poly$(1/\epsilon, \size{P_1}, \dots, \size{P_k})$ steps, where $P_i$ is an input parameter to the procedure. 
It can be readily seen that the approximations and assumptions related to all $b \in B$, and which are required for the algorithm to work like an FPTAS, hold. See \appref{bertfunction} for definitions and proofs of these claims. 

The FPTAS from \newcite{architecturesearch} is an approximation algorithm that relies on optimizing three surrogates to $\mathcal{i}(\cdot)$, $\mathcal{p}(\cdot)$, and $\mathcal{e}(\cdot, \cdot)$, denoted by $\hat{\mathcal{i}}(\cdot)$, $\hat{\mathcal{p}}(\cdot)$, and $\hat{\mathcal{e}}(\cdot, \cdot)$, respectively. This is carried out by expressing them as functions of $\Xi$, and then scalarizing them by selecting a maximum-parameter, maximum-inference-time architecture $T \in B$ called the \emph{maximum point} and via a metric called the $W$\emph{-coefficient}, 

\begin{equation}\label{eq:wcoeffeq}
\mathcal{W}(f, T) = 
\frac{(\mathcal{p}(T) - \mathcal{p}(f))(\hat{\mathcal{i}}(T) - \hat{\mathcal{i}}(f))}
{\mathcal{p}(T)\hat{\mathcal{i}}(T)\hat{\mathcal{e}}(f)}.
\end{equation}
While surrogating $\mathcal{i}(\cdot)$ and $\mathcal{p}(\cdot)$ is relatively straightforward--in fact, $\mathcal{p}(\cdot)\simeq\hat{\mathcal{p}}(\cdot)$--$\mathcal{e}(b(X; W^{*}), Y)$ must be surrogated via the loss function. 
We describe our choices of surrogate functions in the next section. 
Similarly, the guarantees around runtime and approximability depend on two additional input parameters: the chosen number of maximum training steps $n > 0$, as well as the desired interval size $1 \leq \epsilon \leq \size{\Xi}$. The choice of $\epsilon$ directly influences the quality of the solution attained by this approximation algorithm.

\subsection{Setup}\label{sec:fptassetup}

We ran the FPTAS by training our candidate architectures $b \in B$ with the Bookcorpus \cite{bookcorpus} dataset, $n=3$ training steps, interval size $\epsilon = 2$, as well as a fully pre-trained RoBERTa-large as the maximum point $T$. 
This last decision was motivated by the fact that, as mentioned in \secref{relatedwork}, this architecture could be seen as the highest-performing BERT architecture that can be parametrized with some $\xi \in \Xi$. 
Our surrogate error $\hat{\mathcal{e}}(\cdot, \cdot)$ was chosen to be the cross-entropy between the last layers of $T$ and every $b \in B$. Formally, this function is $L$-Lipschitz smooth, but some work is required to show that this property holds with respect to its composition with the BERT architecture, as we show in \appref{xentlip}. The other two conditions--bounded stochastic gradients and boundedness from below--are immediate from the implementation of gradient clipping and finite-precision computation.

For the input, we followed closely the work by \newcite{RoBERTa} and employed an input sequence length of $s = 512$, maintained casing, and utilized the same vocabulary ($V = 50,\!265$) and tokenization procedure. We used stochastic gradient descent as the optimizer, and maintained a single hyperparameter set across all experiments; namely, a batch size of $1,\!024$, learning rate of $5\times10^{-4}$, a gradient clipping constant of $c = 1$, and warmup proportion of $0.05$.

Our search space $\Xi := \mathbf{D}\times\mathbf{A}\times\mathbf{H}\times\mathbf{I}$ consisted of the following architectural parameter sets:  
$\mathbf{D} =\{2,4,6,8,10,12\}$, $\mathbf{A} =\{4,8,12,16\}$, $\mathbf{H} =\{512, 768, 1024\}$, and $\mathbf{I} =\{256, 512, 768, 1024, 3072\}$, corresponding to the depth, attention heads, hidden size, and intermediate size parameter sets, respectively. Note that BERT-base, with $\xi = \langle 12, 12, 768, 3072 \rangle$, is present in this set.\footnote{More precisely, since our embedding layer is based off RoBERTa's, it is RoBERTa-base who is in the search space.} We ignore configurations $\langle D,A,H,I\rangle \in \Xi$ where $H$ is not divisible by $A$, as that would cause implementation issues with the transformer. Refer to \appref{bertmath} for a more detailed explanation of the relationship between these parameters, and \appref{bertfunction} for an explanation of why $\mathbf{D}$ is constrained to even layers. 

We surrogated $\mathcal{i}(\cdot)$ by evaluating the average time that it would take to perform inference on a dataset of $11,\!828$ lines, with the same sequence length, batch size of $1,\!024$, and on a single nVidia V100 GPU. This corpus is simply a small subset of Bookcorpus drawn without substitution, and it is considerably smaller than the original. With a batch size as mentioned above it would still provide a reasonable approximation to the inference latency, and double as our test set for $\hat{\mathcal{e}}(b(X; W^{*}), T(X))$.

\subsection{Results}

We report the top three largest $W$-coefficients resulting from running the FPTAS, along with their parameter sizes, inference speeds, and architectural parameter sets in \tabref{wcoeffs}. 
The candidate architecture with the largest $W$-coefficient (A$1$ in the table), is what we refer to in this work as Bort. Bort was not the smallest architecture, but, as desired, it was the one that provided the best tradeoff between inference speed, parameter size, and the (surrogate) error rate. 
As an additional baseline, we measured the average inference speed on CPU, reported in \tabref{wcoeffs}. %

\begin{table}
\centering
\resizebox{\textwidth}{!}{
\begin{tabular}{ |c||c|c|c||c| } \hline
 Architecture & $\langle D,A,H,I \rangle$ &  $W$-coefficient & Parameters (M) & Inference speed (avg. s/sample) \\ \hline\hline
 \bf{A1 (Bort)} & $\langle 4,	8,	1024, 768 \rangle$	& 8.6 	& 56.14	& 0.308 \\
 A2 			& $\langle 4, 	16,	1024, 512 \rangle$ 	& 7.5 	& 35.20	& 0.314 \\
 A3 			& $\langle 4, 	8,	1024, 512 \rangle$ 	& 6.6 	& 35.20	& 0.318 \\ \hline
 BERT-base 		& $\langle 12, 12, 768, 3072 \rangle$	& 0.7	& 110	& 2.416 \\
 RoBERTa-large  & $\langle 24, 16, 1024,4096 \rangle$	& N/A	& 355	& 6.170  \\ \hline
\end{tabular}}

\caption{Top three results from running the FPTAS from \newcite{architecturesearch} on $\Xi$. We also include other architectures as a comparison, and the results for the experiment on inference speed evaluation. Inference speed is evaluated on an instance with $21$ GiB memory, $8$ Intel Xeon Platinum $8000$ CPUs running at $3.5$ GHz; for a single element of length $s=512$ drawn randomly from a dataset; and averaged across $6,\!250$ trials.}
\label{tab:wcoeffs}
\end{table}

In terms of the characterization of its architectural parameter set, Bort is similar to other compressed variants of the BERT architecture--the most intriguing fact is the depth of the network is $D=4$ for all but one of the models--which provides a good empirical check with respect to our experimental setup. Refer to \tabref{archcomp} for an explicit comparison. 

\begin{table}[h]
\centering
\begin{tabular}{ |c||c|c| } \hline
 Architecture & $\langle D,A,H,I \rangle$ & Parameters (M) \\ \hline\hline
 Bort				 & $\langle  4,  8, 1024, 768\rangle$ 	& 56.1 \\ \hline
 TinyBERT (4 layers) & $\langle  4, 12,  312, 1200\rangle$	& 14.5 \\
 BERT-of-Theseus	 & $\langle  6, 12, 768,  3072\rangle$	& 66 \\ 
 BERT-Mini			 & $\langle  4,  4,  256, 2048\rangle$ 	& 11.3 \\
 BERT-Small 		 & $\langle  4,  8,  512, 2048\rangle$ 	& 29.1 \\ \hline 
 BERT-base 			 & $\langle 12, 12, 768,  3072\rangle$	& 110 \\
 BERT-large 		 & $\langle 24, 16, 1024, 4096\rangle$	& 340 \\ 
 RoBERTa-large  	 & $\langle 24, 16, 1024, 4096\rangle$	& 355 \\ \hline
\end{tabular}
\caption{Architectural comparison between various members of the codomain of $\Xi$. Parameter sizes for other models are as reported by their authors.}
\label{tab:archcomp}
\end{table}

Since our maximum point and vocabulary are based off RoBERTa's, Bort is more closely related to that specific variation of the BERT architecture. Yet, as mentioned before, mathematically the size of the embedding layer does not have a strong influence on the results from the FPTAS, and it is quite likely that the result would not change dramatically. 
It is important to point out that our verification for the $\mathcal{i}(\cdot)$ component will be hardware-dependent, and it is likely to change values. 

That being said, the parameter size of the embedding layer is given by $VH + SH + 3H$, with $S$ being the token type embedding size--$S=514$ for Bort and RoBERTa, $S=512$ for BERT. This layer alone is $52$ million parameters for RoBERTa-large, $31.8$ million parameters for BERT-large, and $39$ million parameters for Bort. This means that the block of encoder layers of Bort, which is the component of the architecture that presents the largest cost in terms of inference speed, is $5.5\%$ the size of BERT-large's, and $5.6\%$ the size of RoBERTa-large's. For a more rigorous analysis on the evaluation of the number of operations on the BERT architecture, refer to \appref{bertmath} and \appref{bertfunction}.

\section{Pre-training Using Knowledge Distillation}\label{sec:distillation}

Although the FPTAS guarantees that we are able to obtain an architectural parameter set that describes an optimal subarchitecture, it still remained an open issue to pre-train the parametrized model efficiently. 
The procedure utilized to pre-train BERT and RoBERTa (self-supervised pre-training) appeared costly and would defy the purposes of an efficient architecture. Yet, pre-training under an intrinsic evaluation metric leads to considerable increases in more specialized, downstream tasks \cite{BERT}. The performance evaluation around said downstream tasks will be the topic of \secref{results}. 
In this section, we discuss our approach, which employed KD, and compare it to self-supervised pre-training, but applied to Bort. 

\subsection{Background}

A common theme on the variants from \secref{relatedwork}, especially the ones by \newcite{tinyBERT} and \newcite{GooglesMiniBERTs}, is that using KD to pre-train such language models led to good performances on the chosen intrinsic evaluation metrics. 
Given that the surrogate error function from \secref{main}, $\hat{\mathcal{e}}(\cdot, \cdot)$, was chosen to be the cross-entropy with respect to the maximum point, it seems natural to extend said evaluation through KD. 

We also compare self-supervised pre-training to KD-based pre-training for the Bort architecture, akin to the work done by \newcite{GooglesMiniBERTs}. We found a straightforward cross-entropy between the last layer of the student and teacher to be sufficient to find a model that resulted in higher masked language model (MLM) accuracy and faster pre-training speed when compared to this other method.

\subsection{Setup}

For both pre-training experiments, and in line with the works by \newcite{BERT} and \newcite{RoBERTa}, we used as a target the MLM accuracy--also known as the Cloze objective \cite{Taylor}--as our primary intrinsic metric. This is mainly due to the coupling between this function and the methodology used for fine-tuning on downstream tasks, where the algorithm iteratively expands the data by introducing perturbations on the training set. 
The MLM accuracy would therefore provide us with an approximate measure of how the model would behave under corpora drawn from different distributions.

In order to have a sufficiently diverse dataset to pre-train Bort, we combined corpora obtained from Wikipedia\footnote{\url{https://www.en.wikipedia.org}; data obtained on September $16^{th}, 2019$}, Wiktionary\footnote{\url{https://www.wiktionary.org}; data obtained in July $3^{rd}$, 2019. We only utilized the definitions longer than ten space-separated words.}, OpenWebText \cite{openwebtext}, UrbanDictionary\footnote{\url{https://www.urbandictionary.com/}; Data corresponding to the entries done between 2011 and 2016. We only utilized the definitions longer than ten space-separated words.}, One Billion Words \cite{chelba2013one}, the news subset of Common Crawl \cite{commoncrawl}\footnote{Following \newcite{RoBERTa}, we crawled and extracted the English news by using news-please \cite{Hamborg2017}. For reproducibility, we limited ourselves to using only the dates (September 2016 and February 2019) as the original RoBERTa paper.}, and Bookcorpus. 
Most of these datasets were used in the pre-training of both RoBERTa and BERT, and are available publicly. We chose the other datasets (namely, UrbanDictionary) because they were not used to pre-train the original models, and hence they would provide a good enough approximation of the generalizability of the teacher for the KD experiment. After removing markup tags and structured columns, we were left with a total of $270$ million sentences of unstructured English text, which was subsequently shuffled and split $8$-$1$-$1$ for train-test-validation. 
Aside from that, our preprocessing followed \newcite{RoBERTa}, rather than \newcite{BERT}, closely: in particular, we employed the same vocabulary and embeddings, generated by Byte-Pair Encoding (BPE) \cite{BPE}, used input sequences of at most $s = 512$ tokens, dynamic masking, and removed the next-sentence prediction training objective. 

For the KD setting, we chose a batch size of $1,024$ with no gradient accumulation, and a total of $8$ nVidia V100 GPUs. 
The relative weight between MLM loss and the distillation loss--that is, the cross-entropy between the teacher and student layers--was set to $0.5$ and the distillation temperature value was set to $2.0$. We used an initial learning rate of $1\times10^{-4}$, scheduled with the so-called Noam's scheduler from \newcite{Attention}, and a gradient clipping of $c = 1$.
Finally, due to the fact that we utilized RoBERTa-large as the maximum point for the FPTAS in \secref{main}, it became the natural choice of teacher in KD pre-training. Some prior work \cite{sun2019patient, Distillbert, tinyBERT} required additional loss terms for the KD setting, or even losses at multiple layers to pre-train the network sufficiently; however, we found the result of only using distillation at the last layer between the student and teacher sufficient, in line with the results by \newcite{GooglesMiniBERTs}. 

For the self-supervised training experiment, we used a batch size of $1,\!024$ on $8$ nVidia V100 GPUs. We ran this training with a gradient clipping of $c = 1$. Additionally, we used a peak learning rate of $1\times 10^{-4}$, which was warmed up over a proportion of $0.01$ steps, and with the same scheduler. 
Prior to launching this experiment, we performed a grid search over a smaller subset of our training data to tune the batch size, learning rate, and warmup proportion. 

\begin{table}[h]
\centering
\begin{tabular}{ |c||c|c||c|c|c| }
\hline
Method & Pre-training & Distillation & BERT-large & RoBERTa-large\\
\hline
Number of GPUs 		& $8 	$	& $8$ 		& $1,\!472$	& $1,\!024$	\\
Training time (hrs) & $40 	$	& $36$ 		& $0.78$	& $24$		\\
GPU hours		 	& $320 	$ 	& $288$ 	& $1,\!153$	& $25,\!764$	\\
MLM accuracy 		& $99.3\%$	& $99.3\%$	& N/A		& N/A		\\
\hline
\end{tabular}
\caption{A comparison between standard (self-supervised) pre-training of Bort versus KD-based pre-training. We observed that the latter converged much faster, and to a better MLM accuracy than its self-supervised counterpart. For reference, we also include the numbers for RoBERTa-large--as reported by the original paper--as well as the current world record for fastest pretraining of BERT-large. All the numbers reported use the same GPU model, but not necessarily the same deep learning framework, or dataset size. }
\label{tab:distill}
\end{table}

\subsection{Results}\label{sec:kdresults}

\begin{figure}
\centering
\includegraphics[scale=0.65]{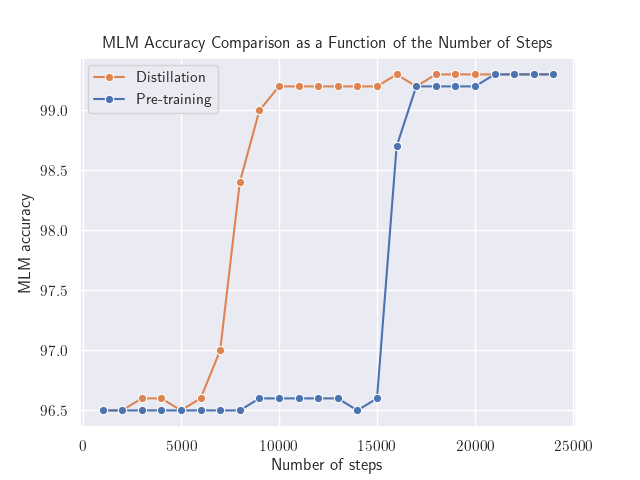}
\caption{MLM performance while pre-training Bort over time, both in the distillation and self-supervised methods. Both approaches were stopped after the $25,\!000$ step mark, and yield the same validation accuracy. The distillation algorithm converges significantly faster, at $17,\!000$ steps versus self-supervised's $22,\!000$.}
\label{fig:kdcomparison}
\end{figure}

While we originally intended to run the self-supervised pre-training until the performance matched BERT's,\footnote{That would be around $98.5\%$ MLM accuracy, as reported in \url{https://github.com/google-research/bert}. Accessed on May $1^{\text{st}}$, 2020.} it eventually became clear that, at least for Bort, KD pre-training was superior to self-supervised training in terms of convergence speed. The results of the comparison are reported in \tabref{distill}, and a plot of their convergence can be found in \figref{kdcomparison}. 

In terms of GPU hours, KD pre-training was considerably more efficient ("greener"), by achieving better results and by using overall $10\%$ less effective cycles needed by its self-supervised counterpart. Although we had a strict cutoff of a maximum of $25,\!000$ training steps, self-supervised pre-training took longer to converge to the same accuracy as KD, with the latter converging at the $17,\!000$ step mark, as opposed to the former's $22,\!000$ step. The number of cycles on the KD scenario also includes the time required to perform inference with the teacher model. 

Compared to the training time for BERT-large ($1,153$ GPU hours for the world record on the same hardware,\footnote{\url{https://developer.nvidia.com/blog/training-bert-with-gpus/}, accessed July $30^{th}, 2020$.} but with a dataset ten times smaller), and RoBERTa-large ($25,764$ hours, with a slightly larger dataset), Bort remains more efficient. We must point out that this comparison is inexact, as the deep learning frameworks for the training of these models are different, although the same GPU model was used across the board. 

These results are not surprising, since the FPTAS selects models based on their convergence speed \cite{architecturesearch} to a stationary point, and ties this stationarity to performance on the training objective. Still, the surrogate loss was chosen to be the cross-entropy, and a series of experiments around extrinsic benchmarks was needed to guarantee generalizability. That being said, the KD-pre-trained Bort became our model of choice for the initialization of the fine-tuning experiments in \secref{results}. 

\section{Evaluation}\label{sec:results}

To verify whether the remarkable generalizability of BERT and RoBERTa were conserved through the optimal subarchitecture extraction process, we fine-tuned Bort on the GLUE and SuperGLUE benchmarks, as well as the RACE dataset. 
The results are considerably better than other BERT-like compressed models, outperforming them by a wide margin on a variety of tasks.

Early on in the fine-tuning process it became patent that Bort was notoriously hard to train, and it was very prone to overfitting. This can be explained by the fact that it is a rather small model, and the optimizer that we used (AdamW) is well-known to overfit without proper tuning being done ahead of time \cite{AdamW}. Likewise, aggressive learning rates and warmup schedules were counterproductive in this case.

While employing teacher-student distillation techinques akin to the ones used in \secref{distillation} showed some promise and brought Bort to a competitive place amongst the other compressed architectures, we noticed that, given its parameter and vocabulary sizes, it could theoretically be able to learn the correct distribution on the input tasks. In particular, small datasets like the Corpus of Linguistic Acceptability (CoLA) by \newcite{cola} should pose no problem to a model with a capacity like Bort's.\footnote{This statement could only be true when measuring such a model with simple accuracy. Matthew's Correlation (the default metric for CoLA) is considerably less forgiving of false positives/negatives, and, indeed, it is a better reflection of a model's ability to understand basic linguistic patterns. In the remaining of this section, we show that our conjecture about the learnability of a dataset with respect to the parameter size was correct for most tasks--with the glaring exception of CoLA.}

In a recent paper, \newcite{finetuningalgo} described a greedy algorithm, known as Agora, that, by combining data augmentation with teacher-student distillation, is able to--under certain conditions--provably converge w.h.p. to the teacher's performance on a given binary classification task. This procedure trains a model $\mathcal{T}$, referred to in the paper as \emph{Timaeus}, on a dataset $X$, and relies on a teacher (\emph{Socrates}) $\mathcal{S}$, a transformation function called the \emph{$\tau$-function}, as well as a starting hyperparameter set $\Theta$. Agora alternatively trains Timaeus, prunes members of $\Theta$, and expands the dataset by using Socrates. As long as the validation data is representative in a certain sense of the task being learned, and the labels are balanced, the algorithm is capable of returning a high performing model with very little data being given initially. 

We employ several RoBERTa-large models as our task-specific Socrates, fine-tuned on every task as prescribed on the original paper. We obtained slightly different oracle performances as compared to the ones described in \tabref{glueresults}, \tabref{superglueresults}, and \tabref{raceresults}, although well above the $2/3$-accuracy requirement from \newcite{finetuningalgo}. 
Additionally, we used as a starting hyperparameter set $\Theta = \mathbf{L}\times\mathbf{D}\times\mathbf{W}\times\mathbf{S}$, 
for learning rates $\mathbf{L} = \{3\times10^{-6}, 5\times10^{-6}, 9\times10^{-6}, 3\times10^{-5}, 1\times10^{-4}\}$, 
weight decays $\mathbf{D} = \{0, 10, 100, 350\}$, 
warmup proportions $\mathbf{W} = \{0.35, 0.40, 0.45, 0.50, 0.55\}$, 
and random seeds $\mathbf{S} = \{0,1,2,3,4\}$. 

It is important to point out that Agora was mainly designed for binary classification problems, and its convergence guarantees are constrained to using accuracy as a metric. While a large portion of the tasks we evaluated fall into that category, we had to modify this procedure to fit regression, question-answering, and multi-class classification problems, with varying degrees of success. We describe the changes made to the problems in their corresponding sections, but a common pattern to all of them was that sometimes we needed to repeat certain iterations of Agora before proceeding to the next, in order to guarantee a much higher-performing architecture.

\subsection{GLUE}\label{sec:glueres}
The Generalized Language Evaluation benchmark is a set of multiple common natural language tasks, mainly focused on natural language inference (NLI). It is comprised of ten datasets: the Stanford Sentiment Treebank (SST-2) \cite{sst2}, the Question NLI (QNLI) \cite{GLUE} dataset, the Quora Question Pairs (QQP)\footnote{The dataset is available at \url{https://www.quora.com/q/quoradata/First-Quora-Dataset-Release-Question-Pairs}} corpus, the MultiNLI Matched and Mismatched (MNLI-m, MNLI-mm) \cite{mnli} tasks, the Semantic Textual Similarity Benchmark (STS-B) \cite{stsb}, the Microsoft Research Paraphrase Corpus (MRPC) \cite{mrpc}, and the Winograd NLI (WNLI) \cite{levesque2011winograd} task. It also includes the Recognizing Textual Entailment (RTE) corpus, which is a concatenation by \newcite{GLUE} of multiple datasets generated by \newcite{dagan2006pascal}, \newcite{bar2006second}, \newcite{giampiccolo2007third}, and \newcite{bentivogli2009fifth}, as well as the aforementioned CoLA, and AX, a diagnostics dataset compiled by \newcite{GLUE}. 

We fine-tuned Bort by adding a single-layer linear classifier in all tasks, with the exception of CoLA, where we noticed that adding an extra linear layer between Bort and the classifier improved convergence speed. All tasks were fine-tuned by using Agora, and we report the full convergent hyperparameter sets in \appref{convparams}.

We used a batch size of $8$ for all tasks with the exception of the four largest ones: SST-2, QNLI, QQP, and MNLI, where we used $16$. Given Bort's propensity to overfit, we opted to initialize our classifier uniformly, drawn from a uniform probability distribution $U(-\frac{1}{2}, \frac{1}{2})$. For all experiments, we maintained a dropout of $0.1$ across all layers of the classifier. Finally, we followed \newcite{BERT} and used a standard cross-entropy loss in all classification tasks, save for STS-B, where we used mean squared error as it is a regression problem. 
We modified Agora to evaluate this task in such a way that it would appear to be a classification problem. This was done through comparing the rounded version of the predictions and the labels, as done by \newcite{T5}.

\begin{table}[h]
\centering
\resizebox{\textwidth}{!}{
\begin{tabular}{ |c||c|c|c|c|c||c| } \hline
 Task 			& RoBERTa-large & BERT-large & BERT-of-Theseus & TinyBERT & BERT-small & Bort \\ \hline\hline
 CoLA			& 67.8 		& 60.5 	  	& 47.8		& 43.3 		& 27.8 		& 63.9 \\
 SST-2	   		& 96.7 		& 94.9 	  	& 92.2		& 92.6 		& 89.7 		& 96.2 \\
 MRPC  	  		& 92.3/89.8 & 89.3/85.4 & 87.6/83.2 & 86.4/81.2 & 83.4/76.2 & 94.1/92.3 \\
 STS-B 			& 92.2/91.9 & 87.6/86.5 & 85.6/84.1 & 81.2/79.9 & 78.8/77.0 & 89.2/88.3 \\
 QQP 			& 74.3/90.2 & 72.1/89.3 & 71.6/89.3 & 71.3/89.2 & 68.1/87.0 & 66.0/85.9 \\
 MNLI-matched	& 90.8		& 86.7 	  	& 82.4		& 82.5 		& 77.6 		& 88.1 \\ 
 MNLI-mismatched& 90.2		& 85.9 	  	& 82.1		& 81.8 		& 77.0 		& 87.8 \\
 QNLI 			& 95.4		& 92.7 	  	& 89.6		& 87.7 		& 86.4 		& 92.3 \\
 RTE 			& 88.2		& 70.1 	  	& 66.2		& 62.9 		& 61.8 		& 82.7 \\
 WNLI			& 89.0		& 65.1 	  	& 65.1		& 65.1 		& 62.3 		& 71.2 \\
 AX				& 48.7		& 39.6 	  	& 9.2 		& 33.7 		& 28.6 		& 51.9 \\ \hline
\end{tabular}}
\caption{Results for fine-tuning Bort on the GLUE benchmarks, as well as the scores for BERT-large, RoBERTa-large, and the three closest architectural variations available. Most tasks are evaluated with accuracy, except: CoLA and AX (Matthew's correlation); STS-B (Pearson correlation / Spearman-$\rho$) and QQP and MRPC (F1 / Accuracy). The results for TinyBERT correspond to the $4$-layer variant. Both TinyBERT and BERT-small have other architectural variations, but these are the closest to Bort. Results for BERT-small are taken from their official repository.}
\label{tab:glueresults}
\end{table}

The results can be seen in \tabref{glueresults}.\footnote{Values obtained from \url{https://gluebenchmark.com/leaderboard} and \url{https://github.com/google-research/bert} on May $3^{\text{rd}}$, 2020.} Bort achieved great results in almost all tasks: with the exception of QQP and QNLI, it performed singificantly better than its other BERT-based equivalents, by obtaining improvements of between $0.3\%$ and $31\%$ with respect to BERT-large. We attribute this to the usage of Agora for fine-tuning, since it allowed the model to better learn the target distributions for each task. 

In QNLI and QQP, this approach failed to provide the considerable increases shown in other tasks. 
We hypothesize that the large size of these datasets, as well as the adversarial nature of the latter,\footnote{\url{https://gluebenchmark.com/faq}} contributed to these results. Recall that Agora requires the validation set to be representative of the true, underlying distribution, and it might not have been necessarily the case in these tasks.

The performance increase varies drastically from task to task, and it rarely outperforms RoBERTa-large. In the case of AX, where the strongest results were present, we utilized the Bort fine-tuned with both MNLI tasks. For MRPC, where paraphrasing involves a much larger amount of lexical overlap, and is measured with both F1 and accuracy, Bort presented the second-strongest results. 
On the other hand, more complex tasks such as CoLA and WNLI presented a significant challenge, since they do require that the model presents a deeper understanding of linguistics to be successful.

\subsection{SuperGLUE}\label{sec:superglueres}

SuperGLUE is a set of multiple common natural language tasks. It is comprised of ten datasets: the Words in Context (WiC) \cite{WiC}, Reasoning with Multiple Sentences (MultiRC) \cite{MultiRC}, Boolean Questions (BoolQ) \cite{BoolQ}, Choice of Plausible Alternatives (COPA) \cite{COPA}, Reading Comprehension with Commonsense Reasoning (ReCoRD) \cite{ReCoRD}, as well as the Winograd Schema Challenge (WSC) \cite{levesque2011winograd}, the Commitment Bank (CB) \cite{CB}, and two diagnostic datasets, broad coverage (AX-b) and Winogender (AX-g) \cite{rudinger2018winogender}. It also includes RTE, as described in the previous section.

\begin{table}[h]
\centering
\begin{tabular}{ |c||c|c|c||c| } \hline
Task    & BERT-large& Baselines & RoBERTa-large & Bort \\ \hline\hline
BoolQ 	& 77.4 		& 79.0 		& 87.1 		& 83.7 		\\
CB 		& 75.7/83.6 & 84.8/90.4 & 90.5/95.2 & 81.9/86.5 \\
COPA 	& 70.6 		& 73.8 		& 90.6 		& 89.6 		\\
MultiRC & 70.0/24.1 & 70.0/24.1 & 84.4/52.5 & 83.7/54.1 \\
ReCoRD 	& 72.0/71.3 & 72.0/71.3 & 90.6/90.0 & 49.8/49.0 \\
RTE 	& 71.7 		& 79.0 		& 88.2 		& 81.2 		\\ 
WiC		& 69.6 		& 69.6 		& 69.9 		& 70.1 		\\
WSC 	& 64.4 		& 64.4 		& 89.0 		& 65.8 		\\
AX-b 	& 23.0 		& 38.0 		& 57.9 		& 48.0 		\\
AX-g  	& 97.8/51.7 & 99.4/51.4 & 91.0/78.1 & 96.1/61.5 \\\hline
\end{tabular}
\caption{Results for fine-tuning Bort on the SuperGLUE benchmarks, as well as the BERT-large and RoBERTa-large scores. The baselines are obtained by training BERT further on related tasks \cite{SuperGLUE}. No results for compressed variants are available at the time of writing this. Most tasks are evaluated with accuracy, except: AX-b (Matthew's correlation); AX-g (Gender Parity/Accuracy); MultiRC (F1a/EM); and CB and ReCoRD (F1 / Accuracy).}
\label{tab:superglueresults}
\end{table}

Just as in \secref{glueres}, we fine-tuned Bort by adding a single-layer linear classifier, and running Agora to convergence in all tasks. 
The results can be seen in \tabref{superglueresults}.\footnote{Data for the other models obtained from \url{https://super.gluebenchmark.com/leaderboard} on May $3^{\text{rd}}$, 2020.} 
The RTE, AX-b and WSC tasks from SuperGLUE are exactly the same as GLUE's RTE, AX, and WNLI, respectively. The latter incorporates information not previously used in its GLUE counterpart: the spans to which the hypothesis refers to in the premise. Likewise, AX-b has the neutral label collapsed into the negative label, thus turning it into a binary classification problem. Bort obtained good results and outperformed or matched BERT-large in all but one task, ReCoRD. 

We re-encoded WNLI to incorporate the span information with a specialized token, which lead to a drop in performance with respect to its GLUE counterpart. We hypothesize the model focused too much on the words delimited by this token, and failed to capture the rest of the information in the sentence. 
Similarly, we noticed that MultiRC had a considerable amount of typos and adversarial examples in all sets. 
This included questions were asked that were not related to the text, ill-posed questions (e.g., "After the Irish kings united, when did was Irish held in Dublin?"), and ambiguous answers with either partial information or no clear referent. To mitigate this, we performed a minor cleaning operation prior to training, removing the typos. 
This task, as well as ReCoRD, is designed to evaluate question-answering. To maintain uniformity accross all our experiments, we casted them into a classification problem by matching correct and incorrect question-answer pairs, and labeling them appropriately. 
Since Agora requires a balanced dataset for some of the results, we artificially induced a balanced dataset by randomly discarding parts of the training set, and enforcing equal proportion of labels in the algorithm's run. We suspect that this could have been detrimental on ReCoRD, where formulating it as a classification problem induced a considerable label imbalance, with a large amount of negative samples distinct from the positive sample by a single word.

\subsection{RACE}
The Reading Comprehension from Examinations (RACE) dataset is a multiple-choice question-answering dataset that emphasizes reading comprehension across multiple NLU tasks (e.g. paraphrasing, summarization, single and multi-sentence reasoning, etcetera), and is collected from the English examinations for high school and middle school Chinese students. 
It is expertly-annotated, and split in two datasets: RACE-H, mined from exams for high school students, and RACE-M, corresponding to middle school tests.
The results can be seen in \tabref{raceresults}.\footnote{Data for the other models obtained from \url{http://www.qizhexie.com/data/RACE_leaderboard.html} on July $30^{\text{th}}$, 2020.}

\begin{table}[h]
\centering
{
\begin{tabular}{ |c||c|c|c|c| } \hline
Task 			& RoBERTa-large & BERT-large & Bort \\ \hline\hline
RACE-M		 	& 86.5 			& 76.6 		 & 85.9  \\\hline
RACE-H		 	& 81.8 			& 70.1 		 & 80.7  \\\hline
\end{tabular}}
\caption{Results for fine-tuning Bort on the RACE dataset, as well as the RoBERTa-large and BERT-large scores. The latter relies on a different training procedure, as reported by \newcite{pan2019improving}. }
\label{tab:raceresults}
\end{table}
As in the previous experiments, we fine-tuned Bort by adding a single-layer linear classifier, and running Agora to convergence. 
Similar to the MultiRC task from \secref{superglueres}, we turned this problem into a classification problem by expanding the correct and incorrect answers and labeling them appropriately. This results in a $3:1$ label imbalance ratio, which was counteracted by randomly discarding about $2/3$ of all the negative samples. Unlike SuperGLUE's ReCoRD, this appeared to not be detrimental to the overall performance of Bort, which we attribute to the fact that negative samples are more easily distinguishable than in the other task. 
Given that the size of this dataset ($20,799$ passages in RACE-H, $7,144$ in RACE-M, with about $16$ possible question-answer pairs) combined with our methodology would slow down training considerably, we pruned it by randomly deleting half of the passages in RACE-H, as well as their associated sets of questions. Overall, Bort obtained good results, outperforming BERT-large by between $9-10\%$ on both tasks.

\section{Conclusion}\label{sec:conclusion}

By applying some state-of-the-art algorithmic techniques, we were able to extract the optimal subarchitecture set for the family of BERT-like architectures, as parametrized by their depth, number of attention heads, and sizes of the hidden and intermediate layer. We also showed that this model is smaller, faster and more efficient to pre-train, and able to outperform nearly every other member of the family across a wide variety of NLU tasks. 

It is important to note that, aside from the algorithm from \newcite{architecturesearch}, no optimizations were performed to Bort, It is certain that techniques such as floating point compression and matrix and tensor factorization could speed up this architecture further, or provide a different optimal set for the FPTAS. Future research directions could certainly exploit these facts. 
Likewise, it was mentioned in \secref{kdresults} that the objective function chosen for the FPTAS was designed to mimic the output of the maximum point. Smarter choices of the surrogate error--likely not as generalizable--would exploit the optimality proofs from this algorithm and create a task-specific objective function, thus finding a highly-specialized, but less-extendable, equivalent of Bort. The existence of these architectures is very much guaranteed by the consequences of the No Free Lunch theorem, as stated by \newcite{ShalevUML}.

The dependence of Bort on Agora is evident, given that without this algorithm we were unable to outmatch the results from \secref{results}. The high performance of this model is clearly dependent on our fine-tuning procedure, whose proof of convergence relies on both a balanced and representative dataset, as well as a lower bound with an inital random model. 
The first condition becomes evident with Bort's poor performance on heavily imbalanced and non-representative datasets, such as QQP and our formulation of ReCoRD. We conjecture that the second condition--a random model--might suggest that similar results could be attained without the fine-tuning step, and for any model. We leave that question open for future work.

Finally, we would like to point out that the success of Bort in terms of faster pre-training and efficient fine-tuning would not have been possible without the existence of a highly optimized BERT--the RoBERTa architecture. 
Given the cost associated with training these models, it might be worthwhile investigating whether it is possible to avoid large, highly optimized models, and focus on smaller representations of the data through more rigorous algorithmic techniques.

\section*{Acknowledgments}
Both authors extend their thanks to H. Lin for his help during the experimental stages of the project, as well as Y. Ibrahim, Y. Guo, and V. Khare. 
Additionally, A. de Wynter would like to thank B. d'Iverno, A. Mottini, and Q. Wang for many fruitful discussions that led to the successful conclusion of this work. 

\bibliography{biblio}

\appendix
\section*{Appendices}
\section{A Description of the BERT Architecture}\label{app:bertmath}
In this section we describe the BERT architecture in detail from a mathematical standpoint, with the aim of building the background needed for the proofs in \appref{bertfunction} and \appref{xentlip}. 
Let $s$ be the sequence length (i.e., dimensionality) for an integer-valued vector $x$, such that $\forall x_i \in x$, $x_i \in V$, where $V$ is a finite set that we refer to as the \emph{vocabulary}. 
When the sense is clear, for clarity we will use interchangeably $V$ to denote the set and its cardinality. Implementation-wise, vocabularies are lookup tables. 
Finally, let $x \in \mathbb{R}^{n \times m}$ be a matrix. We will write the shorthands $W_{m, n}(x)$ to denote a linear layer $f(x) = Wx + b$, where $W \in \mathbb{R}^{m \times n}$; $R(x)$ to denote a dropout layer where $R(x)_{i,j} = 0$ with some probability $\pi$ ($R(x)_{i,j} = x_{i,j}$ otherwise); and, finally, 
$N(x)$ for a layer normalization operation as described by \newcite{layernorm}, of the form $N(x) = \frac{x - \mathbb{E}[x]}{\sqrt{\text{Var}[x] + \epsilon}}\alpha + \beta$, where $\alpha$ and $\beta$ are trainable parameters.\footnote{There are different methods to perform layer normalization, but this is the one that was implemented in the original paper.} 

Then BERT$(x)$ is the composition of three types of functions (input, encoder, and output) that takes in an $s$-dimensional vector $x$,\footnote{In practice, BERT takes in \emph{three} inputs: the integers corresponding to the vocabulary, and the position and token type indices. That can be computed independently and thus we do not consider it part of the architecture.} and returns an $s$-dimensional vector of real numbers,

\begin{equation}\label{eq:fullberteq}
\text{BERT}(x) = \mathbf{p}(\mathbf{l}_{D-1}(\dots(\mathbf{l}_{0}(\mathbf{i}(x)))\dots)),
\end{equation}

with $\mathbf{i}(x)$ being the input, or embedding, layer, defined as:

\begin{equation}\label{eq:inputlayereq}
\mathbf{i}(x) = N(R(V[x] + V'[x'] + V''[x''])). 
\end{equation}

In the input layer, $x'$ and $x''$ are the position (resp. token type) indices corresponding to the input $x$, embedded in its corresponding vocabulary $V'$ (resp. $V''$). 
The BERT architecture has $D$ encoders. Each $\mathbf{l}_i$ encoder is of the form:

\begin{equation}\label{eq:layerlayereq}
\mathbf{l}_i(x) = N(R(W_{I,H}(GeLU(W_{H,I}(\mathbf{m}(x)))) + \mathbf{m}(x) )).
\end{equation}

Each $\mathbf{m}$ is shorthand for:

\begin{equation}\label{eq:intermlayereq}
\mathbf{m}(x) = N(R(W_{H,H}(R(\mathbf{a}(x)))) + x),
\end{equation}

with $\mathbf{a}(x)$ being the attention layer as described by \newcite{Attention},

\begin{equation}\label{eq:attnlayer}
\mathbf{a}(x) = \left(\mathbf{s}\left(\frac{W_{H, H}(x)W_{H,H}^{\top}(x)}{\sqrt{\frac{H}{A}}}\right)\right)W_{H,H}(x).
\end{equation}

In the attention layer, $\mathbf{s}(\cdot)$ is the softmax function, and each of the $W_{H,H}$ correspond to the so-called query, value, and key linear layers. In terms of implementation, since the term $\sqrt{(H/A)}$ denotes the step size with which to apply $\mathbf{s}(\cdot)$, $H$ must be constrained to be divisible by a positive-valued $A$. 

Finally, $\mathbf{p}(x)$ is the output, or pooler, layer:

\begin{equation}\label{eq:outputlayereq}
\mathbf{p}(x) = \tanh{(W_{H,H}(x))}.
\end{equation}

\section{The Weak $AB^nC$ Property in BERT Architectures}\label{app:bertfunction}
In this section we begin to prove the assumptions that allow us to assume that the optimization problem from \secref{main} is solvable through the FPTAS described by \newcite{architecturesearch}. 
Specifically, we show that the family of BERT architectures parametrized as described in \eqref{berteq} presents what is referred to in the paper as the \emph{weak $AB^nC$ property}. 
The existence of such property is a necessary, but not sufficient, condition to warrant that the aforementioned approximation algorithm returns an $(1 -\epsilon)$-approximate solution. The \emph{strong $AB^nC$ property} requires the weak $AB^nC$ property to hold, in addition to conditions around continuity and $L$-Lipschitz smoothness of the loss. These two conditions are be the topic of \appref{xentlip}.

The weak $AB^nC$ property states, informally, that for a neural network of the form $f(x) = C(B_{n -1}(B_{n-2}(\dots B_{0}(A(x)))))$, for any $n \geq 1$, the parameter size and inference speed on some model of computation required to compute the output for the $A(\cdot)$ and $C(\cdot)$ layers are tightly bounded by any of the $B_i(\cdot)$ layers. If there is an architecture $b(\cdot; \cdot)$ with the weak $AB^nC$ property in $B$, then all architectures present such a property \cite{architecturesearch}.

It is comparatively easy to show that BERT presents the weak $AB^nC$ property. To do this, first we state the dependency of $\mathcal{p}(\cdot)$ and $\mathcal{i}(\cdot)$ as polynomials on the variable set for $\Xi$. 
While for the first function it is straightforward to prove said statement via a counting argument, the second depends--formally--on the asymptotic complexity of the operations performed at each layer. 
Instead we formulate $\mathcal{i}(\cdot)$ as a function of $\Xi$ via the number of add-multiply operations (FLOPs) used per-layer. It is important to highlight that, formally, the number of FLOPs is not a correct approximation of the total number of steps required by a mechanical process to output a result. However, it is commonly used in the deep learning literature (see, for example, \newcite{FLOPS}), and, more importantly, the large majority of the operations performed in any $b\in B$ belong to linear layers, and hence, are add-multiply operations. 

To begin, let us approximate the FLOPs required for a linear layer $f(x) = Wx + b$, where $W \in \mathbb{R}^{m \times n}$ as:

\begin{equation}\label{eq:flopseqdef}
[f] = (2n-1)m,
\end{equation}

and denote the number of parameters for the same function as:

\begin{equation}\label{eq:paramseqdef}
\mathcal{p}(f) = mn + n.
\end{equation}

We will make the assumption that the layer normalization and dropout operators are "rolled in", that is, that they can be computed at the same time as a linear layer, so long as said linear layer precedes them. It is common for deep learning frameworks to implement these operations as described.

Similarly, we neglect the computation time for all activation units (e.g., the softmax in \eqref{attnlayer}), with the exception of GeLU. This is due to the fact that the number of FLOPs required to compute an activation unit is smaller than the number of FLOPs for every other component on a network. Given that GeLU is a relatively non-standard function, and most implementations at the time of writing this are hard-coded, we assume that no processor-level optimizations are performed to it. To compute the FLOPs properly, we follow the approximation of \newcite{GeLU},

\begin{equation}\label{eq:geluapprox}
\text{GeLU}(x) = \frac{x}{2}\left(1 + \tanh\left(\sqrt{\frac{2}{\pi}}(x + kx^3)\right)\right),
\end{equation}

where $k = 0.44715$ is a constant. We also note that both \eqref{geluapprox} and its derivative are continuous functions.

\begin{lemma}\label{lem:bertflops}
Let $b$ be a member of the codomain of $\Xi$. The number of FLOPs for $b$ is given by:
\begin{equation}\label{eq:bertflops}
[b] = D(4(2H - 1)H + H^2 + (2H - 1)I + 7I^2) + (2H - 1)H + 3H.
\end{equation}
\end{lemma}
\begin{proof}
By application of \eqref{flopseqdef} to \eqref{fullberteq}. 
\end{proof}

\begin{lemma}\label{lem:bertparams}
Let $b$ be a member of the codomain of $\Xi$. The parameter size for $b$ is given by:
\begin{equation}\label{eq:bertparams}
\mathcal{p}(b) = D(4H^{2} +2HI + 9H + I) + H^2 + (V + S + 6)H.
\end{equation}
\end{lemma}
\begin{proof}
By application of \eqref{paramseqdef} to \eqref{fullberteq}. 
\end{proof}

Note how the dependency of both \eqref{bertflops} and \eqref{bertparams} on $A$ vanishes: on the former, due to the fact that we assumed activation units to be computable in constant time. On the latter, it is mostly an implementation issue: most frameworks encode the attention heads on a single linear transformation. 

\begin{lemma}\label{lem:weakbertprop}
Let

\begin{equation}
b(x; W; \xi) = \mathbf{p}(\mathbf{l}_{D-1}(\dots(\mathbf{l}_{0}(\mathbf{i}(x)))\dots))
\end{equation}

be a BERT architecture parametrized by some $\xi \in \Xi$ such that $b \in B$, and the variable set of $\Xi$ encodes the depth $D$, number of attention heads $A$, hidden size $H$, and intermediate size $I$. If $D$ is even for any $D \in \xi$, then $b(\cdot)$ presents the weak $AB^nC$ property.
\end{lemma}
\begin{proof}
We only provide the proof for the parameter size, as the inference speed follows by a symmetric argument. By \lemref{bertparams}, $\mathcal{p}(\cdot)$ for a single intermediate layer asymptotically bounds the other two layers,

\begin{align}
\mathcal{p}(\mathbf{i}(\cdot)) &\in \BigO{\mathcal{p}(\mathbf{l}_{i}(\cdot))}, \\ 
\mathcal{p}(\mathbf{p}(\cdot)) &\in \BigO{\mathcal{p}(\mathbf{l}_{i}(\cdot))}. 
\end{align}

Given that $D$ is constrained to be even, we can rewrite \eqref{layerlayereq} as:

\begin{equation}\label{eq:layer2eq}
\mathbf{l}_i'(x) = \mathbf{l}_i(\mathbf{l}_{i-1}(x)),
\end{equation}

for $i \in \{ 1, 3, \dots, D-1 \}$. 
The parameter size for \eqref{layer2eq} is a polynomial on the variable set of $\Xi$, and with a maximum degree of $4$ on $H$. Hence, 
\begin{equation}
\mathcal{p}(\mathbf{i}(\cdot; W_{\mathbf{i}}; \xi_{\mathbf{i}})) \in \LittleO{\mathcal{p}(\mathbf{l}_{i}'(\cdot; W_{\mathbf{l}_{i}'}; \xi_{\mathbf{l}_{i}'}))},
\end{equation}
and
\begin{equation}
\mathcal{p}(\mathbf{p}(\cdot; W_{\mathbf{p}}; \xi_{\mathbf{p}})) \in \LittleO{\mathcal{p}(\mathbf{l}_{i}'(\cdot; W_{\mathbf{l}_{i}'}; \xi_{\mathbf{l}_{i}'}))}.
\end{equation}

The proof for the inference speed case would leverage \lemref{bertflops}, and use the fact that $V$ and $S$ are constant.
\end{proof}

\begin{remark}
It is not entirely necessary to have in \lemref{weakbertprop} $D$ to be constrained to even layers. This lemma holds for any geometric progression of the form $D_{i+1} = D_{i} + 2$. 
\end{remark}

\section{The Strong $AB^nC$ Property in BERT Architectures}\label{app:xentlip}

It was stated in \appref{bertfunction} and in \secref{fptasexplanation} that the algorithm from \newcite{architecturesearch} will only behave as an FPTAS if the input presents the strong $AB^nC$ property. Unlike the weak $AB^nC$ property, the strong $AB^nC$ property imposes extra constraints on the surrogate error function, as well as in the architectures, in order to guarantee polynomial-time approximability. These conditions require that the function $\hat{\mathcal{e}}(\cdot,\cdot)$ be $L$-Lipschitz smooth with respect to $W$, bounded from below, and with bounded stochastic gradients. 
The last two constraints can be maintained naturally through the programmatic implementation of a solution (e.g., via gradient clipping),\footnote{Programmatic implementations aside, \newcite{Hahn} proved them to be naturally bounded regardless.} but the first condition--$L$-Lipschitz smoothness of $\hat{\mathcal{e}}(\cdot,\cdot)$--is not straightforward from a computational point of view. 

While our choice of loss function, the cross-entropy between $b(\cdot)$ and $T(\cdot)$, is easily shown to be $L$-Lipschitz smooth if the inputs are themselves $L$-Lipschitz smooth, it is not immediately clear whether $b \in B$ presents such a property.

\begin{theorem}\label{thm:bertllip}
Let $D \geq 1$ be an integer, and let $b \in B$ be of the form

\begin{equation}
b(\cdot; W; \xi) = \mathbf{p}(\mathbf{l}_{D-1}(\dots(\mathbf{l}_{0}(\mathbf{i}(x)))\dots)),
\end{equation}

such that for any weight assignment $W$, $W$ is a closed subset of some compact space $(S, \size{\size{\cdot}})$, where $\min{\size{\size{W}}} = c$, for some $c \geq 1$. Then $b$ is $L$-Lipschitz smooth, with $L \geq c^{10D + 1}$.
\end{theorem}

\begin{proof}

By inspection, \eqsthreeref{inputlayereq}{layerlayereq}{outputlayereq} are all continuously differentiable, and therefore are $L'$-Lipschitz smooth for some $L'$, with respect to any weight assignment $W$. 
Therefore, given that a composition of $m$ $L_1,L_2,\dots,L_m$-Lipschitz smooth functions is $(\prod_{i=1}^m L_i)$-Lipschitz smooth, it follows that $b\in B$ presents such a property. 

The domain of $b : V\times_1\dots\times_sV \rightarrow [-1,1]$ is itself a compact set--the vocabulary, $V \subset \mathbb{N}_{\geq 0}$, is a finite, ordered, integer set--and, from above, it is continuously differentiable. Hence, $b(\cdot)$ is $L$-Lipschitz smooth with respect to both $W$ and its input.

Our argument has only one potential pitfall: \eqref{attnlayer} has a softmax function, which, although continuously differentiable, is not necessarily $L'$-Lipschitz smooth--it maps to $(0,1]$. 
It is straightforward to see that this does not hold when its input $X$ is compact, and it then it reduces to showing that $X$ in every layer of $b(\cdot)$ maintains this property. 
To see this, recall that a finite sequence of $k$ affine transforms, all belonging to some $\mathbb{F}^{m \times n}$, $\tilde{W}_k(\dots(\tilde{W}_1x)\dots)$, is $\size{\size{\tilde{W}}}_p^{k}$-Lipschitz smooth for some $p$. 
Given that $W$ is closed, the outputs of \eqref{inputlayereq} and \eqref{layerlayereq} via the layer normalization operation are then the outputs of some $\size{\size{\tilde{W}}}^k$-Lipschitz smooth function defined over a compact set, shifted and scaled to a zero-mean, unit-variance, likewise compact, set. 
It immediately follows that $\mathbf{s}(x)$ is defined over a compact set, and hence both this function and the attention layer are $L'$-Lipschitz.

The lower bound in $L$ comes from the fact that $\tanh(\cdot)$ is $1$-Lipschitz smooth. Since $\size{\size{W}} \geq c$, and there are $10D + 1$ linear layers present in $b(\cdot)$, the result follows.

\end{proof}

\begin{remark}
Notice that the use of gradient clipping during training implies the existence of such a compact set. Likewise, the lower bound on $L$ is rather loose.
\end{remark}

\begin{corollary}\label{cor:abnc}
Let $\Xi$ be the search space for an instance $I_{BERT}$ of the optimal subarchitecture extraction problem for BERT architectures of the form:

\begin{equation}
b(x; W; \xi) = \mathbf{p}(\mathbf{l}_{D-1}(\dots(\mathbf{l}_{0}(\mathbf{i}(x)))\dots)),
\end{equation}

where $b \in B$ is parametrized by some $\xi \in \Xi$ whose variable set encodes the depth $D$, number of attention heads $A$, hidden size $H$, and intermediate size $I$. If for any $D \in \xi$, for all $\xi \in \Xi$, $D$ is even, the possible weight assignments $W$ belong to a compact set, and the surrogate error function is $L$-Lipschitz smooth, then $I_{BERT}$ presents the strong $AB^nC$ property.
\end{corollary}
\begin{proof}
Follows immediately from \lemref{weakbertprop}, \thmref{bertllip}, and the definition of the strong $AB^nC$ property.
\end{proof}

\section{Convergent Hyperparameter Sets}\label{app:convparams}

As opposed to standard fine-tuning, Agora is a relatively expensive algorithm: for convergence, it requires $\BigO{\size{\Theta}^2}$ full training iterations of the Timaeus model, and the same number of calls to the Socrates model, with a constantly increasing dataset size. 
On the other hand, it is shown in \newcite{finetuningalgo} that the optimal convergent hyperparameter set (or CHS; the final hyperparameter set in Agora's run) is, in theory, the same across every iteration of the algorithm. 
It then follows that, for reproducibility purposes, it is not necessary to run Agora with a search across the entire set $\Theta$, but repeatedly and only over the resulting CHS. 

We include in this section CHS values for every task of the GLUE benchmarks (\tabref{convparamglue}), SuperGLUE benchmarks (\tabref{convparamsuperglue}), and RACE (\tabref{convparamsrace}). 

In all tables, the following nomenclature is used: $lr$ (learning rate), $d$ (weight decay), $wp$ (warmup proportion), $s$ (random seed). The random seed might be implemented differently on other deep learning frameworks, and therefore it is not necessarily transferrable across implementations.

\begin{table}
\centering
\begin{tabular}{ |c||c|c|c|c|c|c| } \hline
 Task & CHS $\langle lr, d, wp, s \rangle$ & Batch size \\ \hline\hline
 CoLA			& $\langle 9\times10^{-6}, 10, 0.45, 3 \rangle$ & $8$  \\
 SST-2	   		& $\langle 1\times10^{-5}, 10, 0.50, 0 \rangle$ & $16$ \\
 MRPC 			& $\langle 1\times10^{-4},100, 0.50, 2 \rangle$ & $8 $ \\
 STS-B 			& $\langle 9\times10^{-6}, 10, 0.50, 1 \rangle$ & $8 $ \\
 QQP 			& $\langle 5\times10^{-5},  0, 0.50, 2 \rangle$ & $16$ \\
 MNLI-matched 	& $\langle 5\times10{-5}, 0.1, 0.50, 3 \rangle$ & $16$ \\ 
 MNLI-mismatched& $\langle 5\times10{-5}, 0.1, 0.45, 0 \rangle$ & $16$ \\
 QNLI 			& $\langle 9\times10^{-6},  0, 0.50, 0 \rangle$ & $16$ \\
 RTE 			& $\langle 1\times10^{-5},100, 0.50, 2 \rangle$ & $8 $ \\
 WNLI			& $\langle 1\times10^{-4}, 10, 0.40, 3 \rangle$ & $8 $ \\
 AX				& $\langle 5\times10{-5}, 0.1, 0.45, 0 \rangle$ & $8 $ \\ \hline
 \end{tabular}
 \caption{Convergent hyperparameter sets for our evaluation of Bort with Agora on the GLUE benchmarks.}
 \label{tab:convparamglue}
\end{table}

\begin{table}
\centering
\begin{tabular}{ |c||c|c|c|c|c|c| } \hline
 Task & CHS $\langle lr, d, wp, s \rangle$ & Batch size \\ \hline\hline
RACE-H 	& $\langle 5\times10^{-5}, 0.1, 0.45, 2 \rangle$ & $16$ \\ \hline
RACE-M 	& $\langle 5\times10^{-5}, 0.1, 0.35, 3 \rangle$ & $16$ \\ \hline
 \end{tabular}
 \caption{Convergent hyperparameter sets for our evaluation of Bort with Agora on the RACE-H and RACE-M datasets.}
 \label{tab:convparamsrace}
\end{table}

\begin{table}
\centering
\begin{tabular}{ |c||c|c|c|c|c|c| } \hline
 Task & CHS $\langle lr, d, wp, s \rangle$ & Batch size \\ \hline\hline
BoolQ 	& $\langle 5\times10^{-6},100, 0.50, 3 \rangle$ & $16$ \\ 
CB 		& $\langle 1\times10^{-5},100, 0.45, 2 \rangle$ & $16$ \\ 
COPA 	& $\langle 1\times10^{-5},100, 0.35, 1 \rangle$ & $16$ \\ 
MultiRC & $\langle 5\times10^{-6}, 10, 0.50, 0 \rangle$ & $16$ \\ 
ReCoRD 	& $\langle 5\times10^{-5},  0, 0.50, 3 \rangle$ & $16$ \\ 
RTE 	& $\langle 1\times10^{-5}, 10, 0.45, 1 \rangle$ & $16$ \\
WiC		& $\langle 5\times10^{-5}, 10, 0.45, 2 \rangle$ & $16$ \\ 
WSC 	& $\langle 5\times10^{-5}, 10, 0.35, 1 \rangle$ & $16$ \\ 
AX-b 	& $\langle 1\times10^{-5}, 10, 0.45, 1 \rangle$ & $8 $ \\ 
AX-g  	& $\langle 1\times10^{-5}, 10, 0.45, 1 \rangle$ & $8 $ \\ \hline
 \end{tabular}
 \caption{Convergent hyperparameter sets for our evaluation of Bort with Agora on the SuperGLUE benchmarks.}
 \label{tab:convparamsuperglue}
\end{table}

\end{document}